%% file: main.tex
\documentclass[letterpaper, 10pt, conference]{ieeeconf}   
\IEEEoverridecommandlockouts
\usepackage[T1]{fontenc}

\usepackage{amsmath,amssymb,amsfonts}
\usepackage{dsfont}
\usepackage{graphicx}
\usepackage{xcolor}
\usepackage{array}
\usepackage{textcomp}
\usepackage{multirow}
\usepackage{booktabs}
\usepackage{colortbl}
\usepackage{float}
\usepackage{pifont}

\usepackage{caption}
\usepackage{subcaption}
\usepackage{multirow,tabularx}
\usepackage{hyperref}
\usepackage{algorithmic}

\usepackage{tikz}
\usetikzlibrary{arrows.meta,positioning,decorations.pathreplacing,calc,fit,shapes.geometric}

\usepackage{balance}
\let\labelindent\relax   
\usepackage{enumitem}
\usepackage{orcidlink}
\usepackage[capitalise,noabbrev]{cleveref}

\hypersetup{colorlinks=true,linkcolor=blue!60!black,citecolor=green!50!black,urlcolor=blue!70!black}
\graphicspath{{figures/}}
\newcommand{\method}{\textsc{P-PoseMem}}
\newcommand{\dproj}{D_{\mathrm{proj}}}

\title{\LARGE \bf
    \textsc{P-PoseMem}: Projective Semantic Memory for Consistent Language Grounding under Pose-Graph Rewrites
}

\author{
    Ha Sier$^{1}$\,\orcidlink{0009-0000-3617-107X},
    Ali Salmasi$^{1}$\,\orcidlink{0009-0001-3982-9962},
    Mengya Xu$^{1}$,
    Haizhou Zhang$^{1}$\,\orcidlink{0009-0005-1321-8687},
    Jie Lu$^{2}$,\\
    Zhuo Zou$^{2}$\,\orcidlink{0000-0002-8546-1329},
    Xianjia~Yu$^{1}$\,\orcidlink{0000-0002-9042-3730} and
    Tomi~Westerlund$^{1}$\,\orcidlink{0000-0002-1793-2694}%
\thanks{This research is supported by the Research Council of Finland's Digital
Waters (DIWA) flagship (Grant No. 359247) and the DIWA Doctoral
Training Pilot project funded by the Ministry of Education and Culture
(Finland). Corresponding authors: Ha Sier and Xianjia Yu.}%
\thanks{$^{1}$Turku Intelligent Embedded and Robotic Systems (TIERS) Lab,
University of Turku, Turku, Finland.
{\tt\footnotesize \{sierha, asalmas, mengya.xu, hazhan, xianjia.yu, tovewe\}@utu.fi}}%
\thanks{$^{2}$School of Information Science and Technology,
Fudan University, Shanghai, China.
{\tt\footnotesize jlu24@m.fudan.edu.cn, zhuo@fudan.edu.cn}}%
}

\IEEEaftertitletext{%
  \vspace{-4pt}
  \begin{center}
    \includegraphics[width=\textwidth]{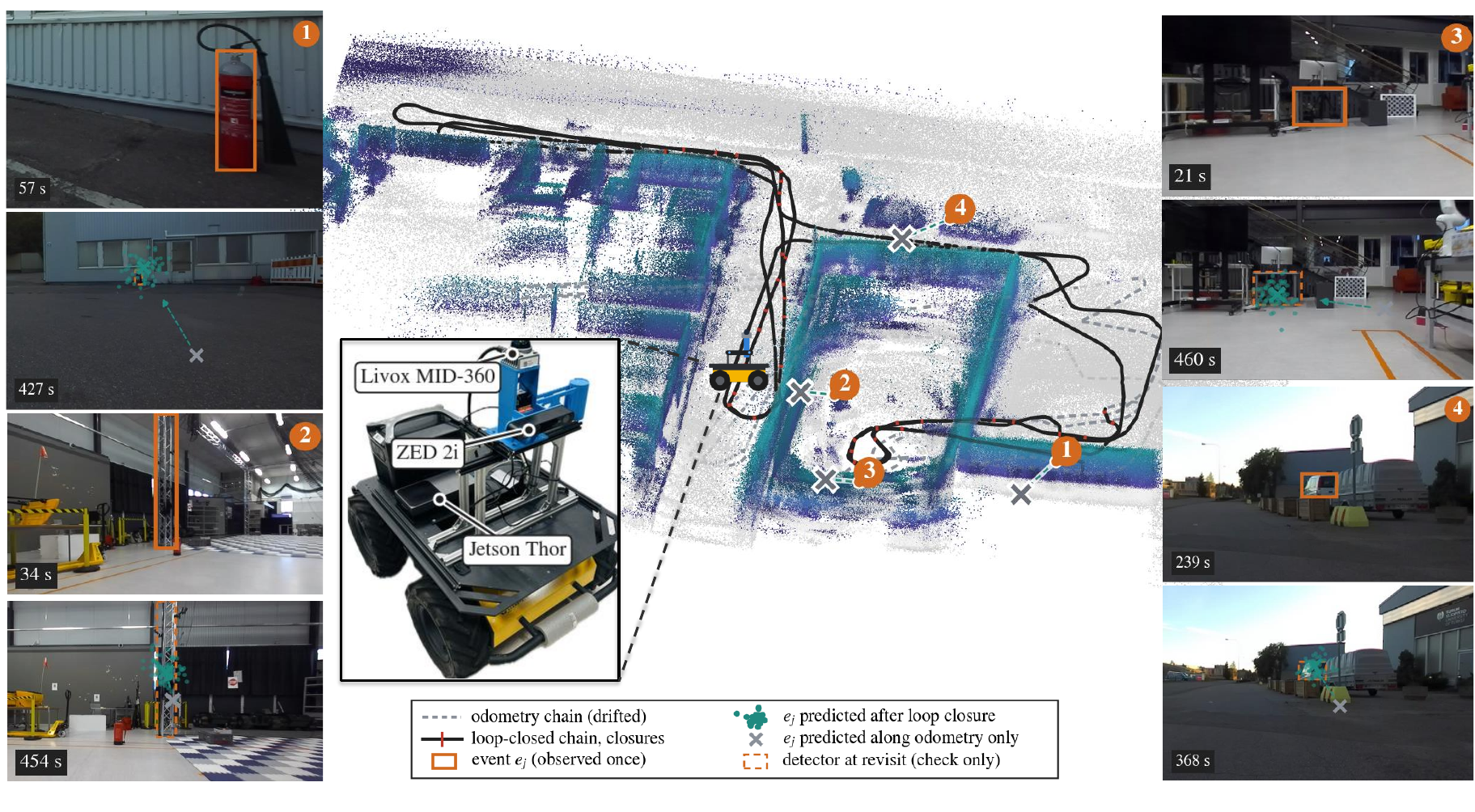}
  \end{center}
  \vspace{-6pt}
  \captionof{figure}{P-PoseMem on one 502\,s drive of the mixed loop. Centre: LiDAR map with the drifted odometry chain (grey), the loop-closed chain (black, closures in red) and events (orange). Around it, four objects seen once (top) and revisited after loop closure (bottom): the memory never re-measures the object, yet its projected posterior lands on it; the grey cross is the odometry-only prediction.}
  \label{fig:teaser}
  \vspace{4pt}
}

\begin{document}
\bstctlcite{IEEEexample:BSTcontrol}

\maketitle

\begin{abstract}
A robot following language instructions needs its semantic memory to keep
naming the same physical object while the SLAM pose graph underneath is
optimized, loop-closed and compressed. Maps committing each detection to a
world coordinate cannot: a closure moves the anchor it was measured from, or
the solver marginalizes that anchor, and the query then selects a different
object although both graphs represent the same posterior.
\method\ stores each observation as an immutable event at its birth keyframe,
retains the Bayes-tree elimination conditional of every marginalized keyframe,
and integrates the semantic likelihood over the reconstructed joint posterior
of poses, anchors and identities. $\dproj$, the total-variation defect between
the language-goal distributions of inference-equivalent full and marginalized
graphs, measures this directly.
Over 40 HM3DSem scenes and $112\,000$ queries \method\ reproduces the
full-graph oracle ($\dproj=0$) and reduces goal flips against every
memory-reducing baseline. On an eight-run campaign whose 761 closures rewrote
the map by up to $47\,\mathrm{m}$, $\dproj$ stays below $10^{-13}$ with
$0/288$ goal flips when elimination follows the closures, where every
ablation and a coordinate committed at insertion flip goals it does not;
under a live bounded solver the same memory flips $23/288$ against $53$ for
that frozen coordinate. A pre-registered negative control is
detected by $\dproj$ while leaving calibration error and navigation success
unchanged, indicating that these measures capture distinct failure modes.
Retrieval is held fixed by a shared frozen detector, isolating the gain to
memory consistency. Code and data: \url{https://anonymous.4open.science/r/posemem-2328/}.
\end{abstract}

\input{sections/introduction}
\input{sections/related_work}
\input{sections/method}
\input{sections/experimental_setup}
\input{sections/results}
\input{sections/discussion}
\input{sections/conclusion}

\bibliographystyle{IEEEtran}
\bibliography{references}

\end{document}

%% file: sections/introduction.tex
\section{Introduction}

A robot told to \emph{go to the fire extinguisher} must
resolve that phrase against a semantic memory built while it was mapping,
and that memory is not static: the pose graph underneath it is optimized on
every keyframe, rewritten whenever a loop closure is accepted, and compressed
whenever the solver marginalizes a keyframe. The same phrase must select the
same physical object before and after each of them, because the person
giving it sees none of them.

Maps that commit each detection to a world coordinate cannot offer that
guarantee. A detection's world position depends on the pose of the
keyframe that measured it, and loop closure moves that pose: on the
campaign of \cref{sec:campaign}, 761 closures moved keyframes by a median
of $2$--$11\,\mathrm{m}$, and a coordinate frozen at birth ends a median
$2.75\,\mathrm{m}$ from its object, over a metre for $86\,\%$ of 1667
re-detected objects. Marginalization is worse: it removes the anchor, and
a memory that kept only a marginal covariance has lost how the removed
anchor depended on the poses that survive. Both are routine solver
operations, and both can silently change which object a query returns.

The components exist and are not joined: open-vocabulary mapping gives
queryable semantic memories~\cite{huang2023vlmaps,gu2024conceptgraphs},
probabilistic grounding gives language likelihoods with
uncertainty~\cite{shao2025morethanapoint,sitdhipol2025fplgn}, and
factor-graph solvers expose the elimination structure that says how a
removed variable depended on those left behind~\cite{kaess2012isam2}.
Missing is a representation in which that structure reaches the query:
language-addressable maps commit geometry at fusion time, scene graphs
corrected under loop closure move a point estimate~\cite{hughes2022hydra},
and uncertainty-aware grounding carries a covariance, not the algebra
that rebuilds an eliminated anchor.

\method\ (Projective PoseMem) makes each semantic observation an immutable
\emph{event}: a local 3D measurement with covariance, a frozen
vision--language embedding and an explicit association hypothesis, in its
birth keyframe's frame and never rewritten. The Gaussian conditional the
solver produces when it eliminates a keyframe is captured verbatim and
linked to the events born there, so a query can draw the retained poses
jointly, draw each removed anchor from its stored conditional, and
integrate the semantic likelihood over the resulting posterior. A graph
update edits posterior and links, never the evidence.

The property is testable, and we test it directly: $\dproj$ is the
total-variation distance between the language-goal distributions of
inference-equivalent full and marginalized graphs under coupled draws, with
the rate at which the arg-max object changes. Storing more numbers will not
drive it to zero; transporting the posterior will, in simulation and on an
eight-run robot campaign alike, where every ablation and a negative control
of the same construction flip goals the memory does not. Our contributions:
\begin{enumerate}[leftmargin=1.3em,itemsep=1pt,topsep=2pt]
  \item An open-vocabulary object memory whose observations are immutable
  events anchored to birth keyframes: graph updates change posterior and
  links, never the evidence.
  \item A memory interface that retains Bayes-tree elimination conditionals
  and rebuilds marginalized anchors at query time, with correlations a
  per-variable covariance discards.
  \item $\dproj$, a graph-rewrite consistency measure on the language-goal
  distribution, not the pose estimate, with ablations and a negative
  control that make it discriminative.
\end{enumerate}

%% file: sections/related_work.tex
\section{Related Work}

Open-vocabulary mapping makes a map queryable by text: VLMaps and
CLIP-Fields fuse vision--language features into a world-frame grid or
field~\cite{huang2023vlmaps,shafiullah2023clipfields}, multimodal spatial
language maps extend the fusion across modalities~\cite{huang2025multimodal},
IVLMap adds per-instance addressing~\cite{huang2025ivlmap}, and
ConceptGraphs and HOV-SG lift the map to object nodes of a scene
graph~\cite{gu2024conceptgraphs,werby2024hovsg}; all commit each entry's
geometry at fusion time, as a function of that moment's pose estimate.
Uncertainty-aware grounding represents where a referent might be rather
than where it is~\cite{shao2025morethanapoint,sitdhipol2025fplgn}, but the
uncertainty stops before the graph: it is not propagated through loop
closure, keyframe elimination and competing associations, nor integrated
over a joint posterior of poses, eliminated anchors and identities.
Object-SLAM places semantic landmarks in the graph, which moves them
correctly by construction, with probabilistic data association keeping
identity soft~\cite{bowman2017probabilistic} or discrete--continuous
smoothing solving assignment and geometry together~\cite{doherty2022dcsam};
but every open-vocabulary detection as a landmark ties the graph to the
detector's output, and the solver returns one assignment. \method\ keeps
a pose graph and the association mixture as a distribution integrated at
query time; none of the above defines a language-goal probability that
survives loop closure and marginalization.

That a closure must move the map is well handled geometrically: Voxgraph
anchors submaps to pose-graph nodes~\cite{reijgwart2020voxgraph}, Hydra
corrects a 3D scene graph under loop closure through a deformation
graph~\cite{hughes2022hydra,hughes2024foundations}, Khronos carries that
into a spatio-temporal map~\cite{schmid2024khronos}, and Clio compresses an
open-set scene graph to what a task needs~\cite{maggio2024clio}. Our
immutable event is the Voxgraph decision at the granularity of one
detection; these systems transport a \emph{point estimate} and do not
model elimination, \method\ transports a \emph{distribution} and
guarantees the language-goal posterior. Clio, closest to our compression
premise, has no equivalence test between the two maps; $\dproj$ is that
test. Marginalization, sparsification and keyframe culling are standard for
bounded-memory
SLAM~\cite{kretzschmar2012compression,carlevarisbianco2014glc,mazuran2014sparsification};
those that resparsify the result trade exactness for a KL-optimal
approximation, while elimination itself is exact for the pose estimate:
eliminating a variable under a
linear-Gaussian model yields a conditional that, retained, reproduces the
joint, and the Bayes tree exposes it~\cite{kaess2012isam2,dellaert2017factorgraphs}.
We build on that theory; our contribution is to make the
conditional structure it produces a first-class part of the semantic
memory, so that language grounding downstream stays invariant to graph
reduction: the conditional is retained, linked to the events born at the
eliminated keyframe, and required to keep the \emph{query
distribution} equivalent, not the pose estimate.

%% file: sections/method.tex
\section{Method}

\begin{figure}[tb]
  \centering
  \includegraphics[width=\columnwidth]{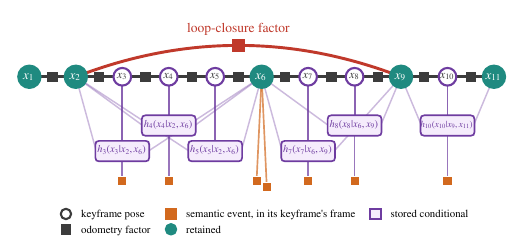}
  \caption{What the memory keeps of the graph: five of eleven keyframes retained (filled), each removed pose $x_k$ leaving its conditional $h_k(x_k \mid x_{\mathcal{S}_k})$ on the survivors, with the events born at $x_k$ pointing at that record.}
  \label{fig:graph}
\end{figure}

\begin{figure*}[t]
  \centering
  \includegraphics[width=\textwidth]{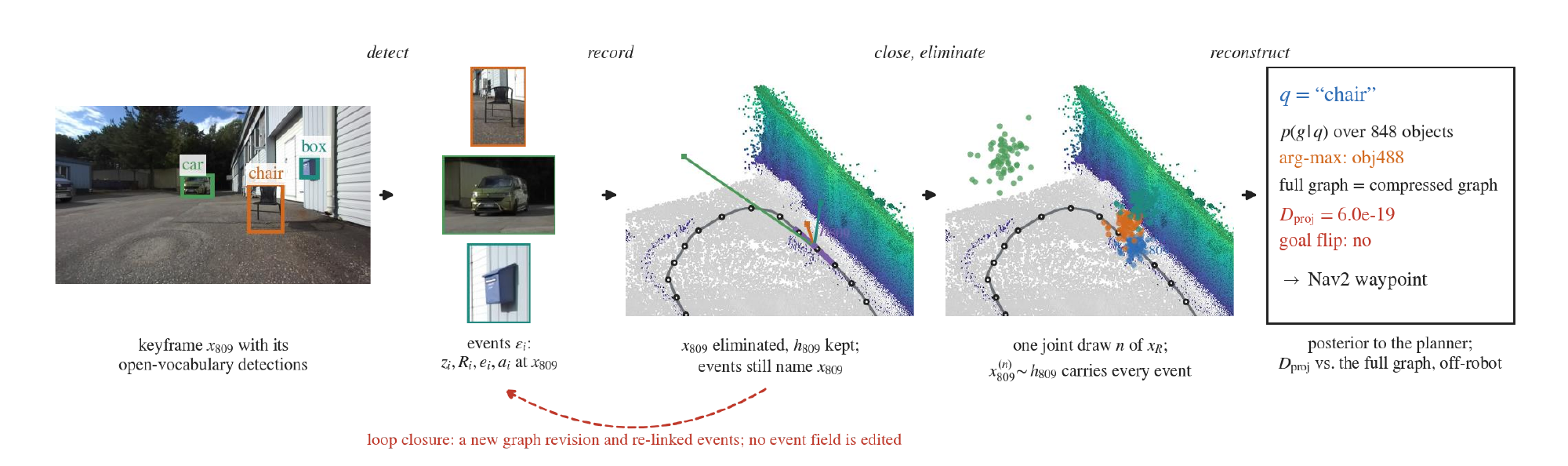}
  \caption{The pipeline on three objects of one keyframe $x_{809}$: detections, the crops the encoder embedded (events $\varepsilon_i$, stored in $x_{809}$'s frame and never edited), their squares on the map, and the 64-draw clouds once $x_{809}$ is eliminated and put back from $h_{809}$ under one joint draw of the retained ones.}
  \label{fig:pipeline}
\end{figure*}

\method\ has three parts: an \emph{event memory} that holds semantic
observations in the frame they were measured from (\cref{sec:m1}), a
\emph{conditional record} of what the solver knows about each keyframe it
removes (\cref{sec:m2}), and a \emph{projective query} that reconstructs
the joint posterior from the two and integrates the language likelihood
over it (\cref{sec:m3}); \cref{sec:dproj} the measure. \Cref{fig:pipeline} shows the three at work
on one lap of the campaign, \cref{fig:graph} the graph they leave behind.

\subsection{Immutable events at their birth keyframe}
\label{sec:m1}
Each detection produces one append-only event: a local position $z_i$
with covariance $R_i$, backprojected from RGB-D into the frame of the
keyframe $x_{k(i)}$ that measured it; a unit-normalized embedding $e_i$ of
the masked crop from a frozen vision--language
encoder~\cite{radford2021clip}; an association hypothesis $a_i$ against the
current object set; and the birth keyframe, timestamp and encoder version.
The three factorize,
\begin{equation}
  p(z_i, e_i, a_i \mid x_{k(i)}, o_{a_i})
  = p(z_i \mid x_{k(i)}, o_{a_i})\, p(e_i \mid o_{a_i})\, p(a_i),
  \label{eq:event}
\end{equation}
so geometry, semantics and identity are each revised by their own
evidence. Identity is a distribution over gated candidates. An
object survives the gate if the observation is inside its $99\,\%$
three-degree-of-freedom Mahalanobis ellipsoid \emph{and} the embeddings
agree above a cosine floor frozen on development scenes; a surviving
candidate carries $\log \pi_a - \tfrac{1}{2}d_a^2 + \kappa\cos_a$, and the
always-present $\mathit{new\_object}$ branch carries $\log \pi_{\mathrm{new}}
+ \kappa$, its cosine being $1$ against its own embedding. The $H=32$
highest-weighted candidates \emph{per observation} are kept and
softmax-normalized, at most one observation per object per keyframe. Local geometry is what makes a later loop closure a
change of \emph{posterior} rather than of \emph{evidence}: a graph
update may add versioned links from an event to a keyframe, never edit a
field.

\subsection{Retaining the elimination conditional}
\label{sec:m2}
Poses, odometry and accepted closures form a factor graph solved
incrementally by iSAM2 in
GTSAM~\cite{kaess2012isam2,dellaert2017factorgraphs}, whose Bayes tree is
retained. When a keyframe $x_k$ is removed, the solver produces a Gaussian
conditional on its \emph{separator} $\mathcal{S}_k$, the surviving poses
it shared factors with,
\begin{equation}
  h_k\!\left(x_k \mid x_{\mathcal{S}_k}\right)
  = \mathcal{N}\!\left(x_k;\, A_k x_{\mathcal{S}_k} + b_k,\ \Sigma_k\right),
  \label{eq:conditional}
\end{equation}
captured verbatim with its separator keys, linearization points,
elimination index and graph revision, and linked by a typed edge to every
event born at $x_k$; a separator eliminated later appends its own
conditional to the chain. Unlike a marginal covariance, the conditional
keeps \emph{how} the removed anchor depends on the survivors, which
decides whether two events correlated before the removal stay correlated
after it.

\subsection{Projective reconstruction and the language goal}
\label{sec:m3}
A query collects every eliminated anchor and surviving separator its
candidate events depend on and processes them in reverse elimination
order: the retained poses are drawn \emph{jointly} from the current Bayes
tree, then each removed keyframe from its stored conditional, one joint
draw reused for every event so that shared uncertainty stays correlated.
The retained conditionals turn the reduced posterior into
\begin{equation}
  p\!\left(o, a, x_R, x_E \mid Z, G'\right)
  = p\!\left(o, a, x_R \mid Z, G'\right)
    \prod_{k \in E} h_k\!\left(x_k \mid x_{\mathcal{S}_k}\right),
  \label{eq:projective}
\end{equation}
with $Z$ the measurements, $G'$ the reduced graph, $o$ the object set,
$a$ the joint assignment of events to objects, $E$ the eliminated anchors
and $R$ the retained ones. The text query is encoded by the same frozen
encoder; each object aggregates its events as $e_g =
\mathrm{norm}(\sum_i w_i e_i)$, $w_i$ proportional to detector confidence
and inverse to $\mathrm{tr}(R_i)$; and the goal posterior
\begin{equation}
  p\!\left(g \mid q, Z, G'\right)
  \propto \sum_a \int
    \exp\!\left(\tau\, \mathrm{sim}(e_q, e_g)\right)
    p\!\left(g, a, x \mid Z, G'\right) dx
  \label{eq:goal}
\end{equation}
is evaluated for the goal $g \in o$ with $\tau$ fitted once on held-out
labelled queries. The sum over $a$ is not enumerated: each pose draw fixes
every observation's world position and covariance and every object's
reliability-weighted state, hence which candidates pass the gate and how the
mixture splits, and the per-observation weights are accumulated into a
per-object mass that is averaged over the draws. Pose draws
enter \cref{eq:goal} through $a$ and $e_g$ alone, deciding which events a
draw lets share an object and so what it looks like; the semantic term
itself is independent of $x$.

\subsection{Measuring graph-rewrite consistency}
\label{sec:dproj}
The quantity that must stay invariant is the distribution the robot acts
on, so we compare the goal distributions of \emph{inference-equivalent}
graphs, the same factors and linearization with and without elimination,
on the union of their object IDs under \emph{coupled} draws, the same
standard-normal residual per variable key on both sides,
\begin{equation}
  \dproj = \tfrac{1}{2} \sum_g
    \left| p_{\mathrm{full}}(g \mid q) - p_{\mathrm{marg}}(g \mid q) \right| ,
  \label{eq:dproj}
\end{equation}
and record whether the arg-max object changes, a \emph{goal flip}. The full-graph side is a verification mirror; the robot
carries only the reduced memory. $\dproj=0$ for \method\ is exact by construction when
both graphs share one linearization, i.e.\ when elimination follows the
closures; a solver that eliminates while closures keep arriving
relinearizes after the conditional was captured, and the reconstruction
inherits that gap (\cref{sec:discussion}). Exactness gives a controlled
verification of the representation; utility is measured separately, by the ablations
of \cref{sec:ablation} and the bounded-solver experiments.

%% file: sections/experimental_setup.tex
\section{Experimental Setup}
\label{sec:campaign}\label{sec:frontend}

\subsection{Perception, simulation and baselines}
Detection and segmentation use a YOLOE open-vocabulary
detector~\cite{wang2025yoloe}; crops are encoded by a frozen MobileCLIP-S1
image encoder~\cite{vasu2024mobileclip} under a fixed crop rule, with
$\tau=60$ fitted on one half of a hand-audited object set and reported on
the other. Every method reads the same frozen, hashed perception cache,
and each query is checked to carry the same configuration, graph revision
and candidate set across methods, so differences between methods are
differences of representation. Simulation scenes come from
HM3DSem~\cite{yadav2023hm3dsem,ramakrishnan2021hm3d} rendered in
Habitat~\cite{savva2019habitat}, thresholds frozen on development scenes
before the test set was opened: 40 scenes $\times$ 16 conditions,
$112\,000$ queries answered through each of the six backends, with injectors for odometry drift, delayed loop
closure, keyframe elimination at $25/50/75\,\%$ and association
difficulty. Baselines are memory reductions of the same pipeline (B0: a world
coordinate committed at fusion time, as VLMaps and CLIP-Fields; B1: the
point estimate re-anchored to the current keyframe pose, as Hydra; B2: a
per-variable marginal covariance, as uncertainty-aware grounding; B3: the
pre-closure snapshot), and B4 is a full-graph oracle that never
eliminates, the reference for $\dproj$. The memory needs a back end that
exposes elimination conditionals: GTSAM's Bayes tree does, g2o and Ceres
stacks do not.

\subsection{Robot campaign, public data and reference}
\label{sec:gt}
The robot is a Clearpath Husky A200 with a ZED 2i stereo camera (the
odometry source) and a Livox MID-360 LiDAR; perception, memory and
retrieval run on board on an NVIDIA Jetson AGX Thor. Loop closures
are \emph{detected}, not asserted: RTAB-Map~\cite{labbe2019rtabmap}
proposes them over the recorded stream and accepted ones enter the graph
as translation-only constraints, the proposals' rotations being
$7.8^\circ$ off the reference at the median. Every campaign graph also carries a legacy last-to-first
odometry factor; removing it moves the B0/B1/B3 flips of \cref{tab:robot}
from $52/22/44$ to $47/26/28$ of $288$ and leaves \method\ at zero, and we
report the graphs as run. Four geometries: two compact
loops of about $50\times40\,\mathrm{m}$ driven three times each, one
crossing between indoors and outdoors; a $184\,\mathrm{m}$ corridor out
and back ($757\,\mathrm{m}$); and an open $150\times85\,\mathrm{m}$ area
($864\,\mathrm{m}$); 24\,632 events over 8274 objects.
OpenLORIS-Scene~\cite{shi2020openloris} adds a wheeled robot driving the
same market and corridor repeatedly with laser-SLAM ground truth; a
scene's sessions are concatenated, each starting at its ground-truth pose
and drifting freely, closures vetted at $1\,\mathrm{m}$. The reference,
LiDAR odometry (FAST-LIO2~\cite{xu2022fastlio2}), closes a
$99.18\,\mathrm{m}$ indoor loop to $0.10\,\%$ against the stereo VIO's
$20.07\,\%$ and vets every accepted closure; a \emph{relative} reference, which
is what an evaluation of graph correction rather than of absolute
localization requires.

\subsection{Queries and navigation}
Consistency is measured on a fixed vocabulary of 36 campaign words against
every run's memory ($288$ queries). Retrieval accuracy, which needs a
hand-labelled target, is scored on twenty audited targets per run on the
six compact-loop runs ($120$ decisions) across difficulty bands by how
many objects share the category (unique; few, 2--5; many, $>5$);
\emph{instance} correctness asks for the exact object, \emph{category}
correctness for the right kind. Planning trials snap each goal to the
nearest traversable cell of a LiDAR occupancy map and call the Nav2
planner~\cite{macenski2020marathon2} at the platform's own footprint
radius and at a conservative one (path efficiency: straight line over
path length).
Execution trials drove the robot on the compact loop to goals from its
own memory for five objects tape-surveyed from the LiDAR, VIO having
drifted $0.2$--$2.8\,\mathrm{m}$ by each first sighting. Identity is fixed
by the survey: the memory object nearest each surveyed position with the
most goal-posterior mass for the category supplies two goals, its
\method\ coordinate and its coordinate frozen at insertion (B0); a goal
inside an inflated obstacle is moved to the nearest standable point and
the move charged to the memory; the score is the final LiDAR pose against
the surveyed object, and \emph{arrival} (within $2\,\mathrm{m}$, object in reach)
was scored at the stop by the operator, unblinded and after the fact.

\paragraph{Reproducibility}
Every number and figure is regenerated by the scripts in the repository,
which also pins the software versions; queries use 64 draws from seed 0.

%% file: sections/results.tex
\section{Results}

\input{tables/tab_rewrites}
\input{tables/tab_robot}
\input{tables/tab_reproj}
\begin{figure}[tb]
  \centering
  \includegraphics[width=\columnwidth]{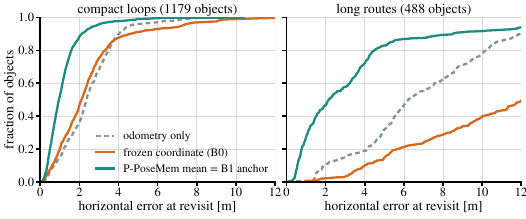}
  \caption{Ground-plane error at revisit on the compact loops and the long routes (per run: \cref{tab:reproj}). The gap between \emph{frozen} and \emph{P} is the representation alone, and it widens with the length of the route.}
  \label{fig:reproj}
\end{figure}

\subsection{Simulation: the oracle and the baselines}
On the analytic gate (a three-pose chain, a ten-pose loop, random sparse
graphs) reconstructed joint means match the full graph to better than
$10^{-8}$ and covariances to $10^{-6}$, with $\dproj=0$ under common draws;
a control that drops the cross-covariance between eliminated and retained
variables fails the same gate at $6\times10^{-4}$. On the 40-scene grid,
across every condition including $25/50/75\,\%$ elimination, drift and
delayed closure, \method's maximum $\dproj$ against the oracle B4 is
$0.000$ with no goal flips, the two agreeing byte-for-byte on all 32
non-key columns of the $112\,000$ paired rows, one per query. This certifies the
implementation (\cref{sec:dproj}); the comparisons that follow carry the
information. On the separate ablation grid the four removals order by how
much structure each discards, mean $\dproj$ $0.010 < 0.022 < 0.043 < 0.086$
for $-$Conditional, $-$Association, $-$Reliability and $-$Correlation.

Under a scene-clustered bootstrap with $10^4$ resamples and Holm
correction over the four baseline contrasts~\cite{holm1979}, \method\ reduces the goal-flip rate against
the oracle for every baseline: $-0.261$ $[-0.328, -0.196]$ vs.\ B1,
$-0.112$ $[-0.145, -0.082]$ vs.\ B2, $-0.255$ $[-0.321, -0.192]$ vs.\ B3
and vs.\ B0, all at Holm-corrected $p \le 4\times10^{-4}$, the resolution
floor of the resampling (the contrasts are scene-weighted bootstrap means,
so they differ from the pooled rates of \cref{tab:robot} by up to $0.02$). $\dproj$ itself is lower than every baseline's,
from $-0.085$ $[-0.092, -0.077]$ (B2) to $-0.159$ $[-0.190, -0.133]$ (B1).

\subsection{The robot campaign}

\Cref{tab:campaign} reports the eight runs over the four geometries of
\cref{sec:campaign}. 761 loop closures were detected, accepted and vetted
against the LiDAR reference, rewriting the graph by a median of
$2$--$11\,\mathrm{m}$ and up to $47\,\mathrm{m}$ on the corridor; $95\,\%$ of
keyframes were eliminated, 64 retained per run. Under that rewrite
\method's $\dproj$ stays below $10^{-13}$ and the arg-max never changes:
$0$ flips of $288$ queries. The same queries from the same events with one part
removed flip between $23$ and $165$ times (\cref{tab:campaign} per
ablation, \cref{tab:robot} for the run-clustered rates,
\cref{sec:ablation} for what each removal isolates). Those are the
replay's counts; the offline rebuild of the same graphs, the source of
every baseline number below, reproduces them within three flips per
ablation.
\label{sec:robotbase}

Answering the same queries through B0--B4 on that rebuild, a coordinate
frozen at insertion (B0) returns a different object from the
oracle on $18\,\%$ of the 288 queries, the pre-closure snapshot (B3) on
$15\,\%$, the current optimum without uncertainty (B1) on $8\,\%$: the
motivating failure is on the robot, at the scale of the closures.
The larger harm is decorrelation: independent per-variable draws (B2,
also the $-$Correlation ablation) flip $58\,\%$. A distribution of the wrong shape moves the arg-max more
often than a misplaced point.

Calibration surprised us: the goal posterior is overconfident for every
backend alike (mean confidence $0.17$ against an instance top-1 of
$0.07$, ECE $0.10$--$0.14$), dominated by the retrieval floor of
\cref{sec:retrieval}, and the negative control is caught by $\dproj$ and
by no other measure (\cref{sec:ablation}).

Elimination bounds the live solver, not the archive. With 64 retained
poses the reduced graph is $0.08\,\mathrm{MB}$ per 1000 keyframes against
$0.67\,\mathrm{MB}$ for the full linearized graph, while the retained
conditionals, which \emph{are} the eliminated part of its Bayes tree, cost
$6.1\,\mathrm{MB}$ under the index order the campaign ran and
$1.6\,\mathrm{MB}$ under a constrained COLAMD order we did not run.
\method\ trades that storage for a closure applied in
$6$--$8\,\mathrm{ms}$ instead of $46$--$165\,\mathrm{ms}$ across the four
geometries. The archive-free alternative, every object a
landmark variable with a bearing--range factor per event and never
eliminated, is exact under pose elimination by construction and fixes
identity at insertion (the $-$Association defect), but keeps every object
in the live solver: on three compact loops ($848$--$1184$ objects) a
closure then costs $88$--$210\,\mathrm{ms}$ over $3270$--$4014$ live
dimensions against $5$--$6\,\mathrm{ms}$ over $462$--$726$ on those same
three runs, slower than the full pose graph ($17$--$27\,\mathrm{ms}$).

\paragraph{The bounded regime, measured}
Every number above is for one elimination after the last closure. We also
replayed each run through the bounded solver itself (64 poses, closures
re-attached to the nearest live keyframe), reconstructed every eliminated
keyframe from its conditional in that conditional's own tangent space, and
compared the goal distributions with the same factor set solved without
elimination (\emph{twin}). \method\ reads a mean $\dproj$ of $0.034$
$[0.024, 0.046]$ against the twin and $0.047$ $[0.032, 0.063]$ against the
batch graph, with $23$ and $26$ flips of $288$; the two references disagree
with each other by $0.027$ $[0.017, 0.038]$ and $17$ flips, an interval
overlapping the first: in the live bounded regime the remaining discrepancy
is on the same scale as the disagreement between the un-eliminated twin and
the batch optimum. A frozen coordinate on the same graph flips $53$, a reconstruction
blind to the relinearization $104$. Geometry pays
more: the retained poses sit $0.7$--$28\,\mathrm{m}$ (per-run medians) from
the twin's optimum and the reconstructed anchors $5$--$26\,\mathrm{m}$,
because the marginal factors kept for the eliminated chains are linear at
the point of capture; re-solving those chains changes nothing, nor does a
four times larger budget. Periodic re-elimination repairs it: once a
retained pose has moved $2\,\mathrm{m}$ from its last elimination
linearization, the solver re-solves the full graph from its current
estimate, re-linearizes and eliminates every archived keyframe again. It
triggers on two closures in five, each costing what one full-graph closure
costs, and brings $\dproj$ to $0.007$ $[0.006, 0.009]$ with $6$ flips, now below
the references' own disagreement, and the anchors to within the twin's
distance from the batch optimum. The goal
distribution survives without the repair because poses reach it through
association alone (\cref{sec:m3}); a navigation goal read off the anchor
needs it.

\Cref{tab:openloris} repeats the measurement on OpenLORIS-Scene, whose
cross-session compositions rewrite the graph by a median of $8\,\mathrm{m}$
(maximum $37$), like the campaign's long routes: $46$ vetted closures,
$\dproj$ at machine precision, $0$ flips of $89$, and the ablations in the
campaign's order at larger size (dropping the conditional or the joint
draw flips $69$ and $68$ of $89$; mean $\dproj$ $0.53$--$0.65$ against
$0.31$--$0.32$). On market 1--2, where the vet refused both closures and the graph was
never corrected, $-$Association flips $10$ of $32$; market 1--3 carries
$3537$ objects through a $25\,\mathrm{m}$ rewrite unchanged.

\subsection{Where the memory puts an object it saw once}
\label{sec:reproj}

\Cref{fig:teaser} shows single objects; \cref{tab:reproj} and
\cref{fig:reproj} report all of them: 1667 objects seen on one pass and
re-detected at least $60\,\mathrm{s}$ and one closure later (80\,864
pairs). Against the LiDAR reference, the birth measurement transported on
the loop-closed chain lands a median $1.11\,\mathrm{m}$ from the object; a
coordinate committed at birth $2.75\,\mathrm{m}$, the raw odometry chain
$2.85\,\mathrm{m}$. That transported point is \method's point prediction
for a once-seen object and, identically, B1's re-anchored coordinate:
every geometric number here is re-anchoring's gain over freezing, which
B1 shares; what \method\ adds over B1 is the distribution around the
point (its coverage, below) and the arg-max consistency of
\cref{tab:robot}, where B1 flips $8\,\%$ and \method\ none.
Run-clustered, the frozen median is $[1.89, 7.31]$ and \method's
$[0.90, 1.40]\,\mathrm{m}$, the paired difference $1.61$ $[0.77, 4.20]$.
The ordering holds on every run and widens with the route, twofold on the
compact loops and fivefold on the corridor and the open area
(\cref{tab:reproj}). \method\ also lands inside the re-detection box more often, though
there the odometry chain overtakes the frozen coordinate
(\cref{tab:reproj}); the residual is the
graph's ($3$--$5^\circ$ of yaw, $7$--$10^\circ$ of pitch, closures
constraining translation only) and stereo ranging's, about one object at
$5$--$10\,\mathrm{m}$.

Calibration of that distribution remains limited by the upstream graph
noise model: the nominal $95\,\%$ ellipsoid covers $33\,\%$ of the pairs ($21\,\%$ mixed, $63\,\%$ open
area), the graph's noise model carrying a median pose spread of
$0.36\,\mathrm{m}$ against a $1.1\,\mathrm{m}$ error. The two properties
separate: rebuilding every graph with its sigmas scaled by $k$ leaves the
optimum in place to the millimetre and raises coverage to $0.60$, $0.79$,
$0.89$ and $0.98$ for $k=2,3,4,8$, the reduced graph still reconstructing
the full graph's draws to $10^{-12}$. \method\ preserves the represented belief
under graph reduction; the calibration of that belief stays determined by
the upstream sensing and graph noise models, here about four times too
narrow.

\subsection{Retrieval and navigation}
\label{sec:retrieval}
Across 120 language decisions, instance top-1 is $6.7\,\%$, top-5
$22.5\,\%$ and category top-1 $57.5\,\%$. The evaluation spans the difficulty bands and most
targets fall in the hardest: $78$ of $120$ lie in the \emph{many} band with
a median of 55 same-category objects, queried by a bare noun. The bound is the shared front end. A human audit of 200
language--target pairs ($183$ judged definitely) found $28\,\%$ of the
definite ones resting on a mislabelled detection (clean rate $0.721$
$[0.638, 0.800]$ scene-clustered, re-audit $\kappa=0.775$), and it concentrates by
detector confidence and by word. Instance
top-1 across backends is $5.8$--$8.3\,\%$ and top-5 $20.8$--$24.2\,\%$
at $n=120$, where the comparison could resolve a
difference of $9$ points: retrieval is statistically indistinguishable
across backends, as their shared frozen front end predicts, and
at that floor most flips do not alter correctness: $58$ of
$-$Conditional's $66$ audited flips exchange one wrong object for another.
\label{sec:navexec}

With retrieval held at that floor, navigation tests the coordinate. Of 12
planning trials $11$ are reachable at the platform's radius and $8$ at
the conservative one (path efficiency $0.91$ and $0.96$; over the ten
drives, success weighted by path length is $0.88$ for the memory against
$0.86$ for the frozen coordinate). In the execution
trials (\cref{sec:campaign}; the \method\ goal is the draw mean, i.e.\
B1's coordinate) all ten drives reached their goal to
$0.07$--$0.33\,\mathrm{m}$; the final pose was $0.27$--$1.43\,\mathrm{m}$
from the object under \method\ (median $0.55$) against
$0.35$--$2.44\,\mathrm{m}$ under the frozen coordinate (median $0.98$).
Distance is not arrival: at one extinguisher the frozen coordinate, off
$0.9\,\mathrm{m}$ on the far side of its partition, left the object out of
reach, the memory's, off $1.0\,\mathrm{m}$ on the near side, in reach. By
the operator's criterion the memory arrives at five of five objects and
the frozen coordinate at three (the bicycle at $2.44\,\mathrm{m}$ is its
other miss); the memory's coordinate is closer on four of five, by
$0.06$--$0.55\,\mathrm{m}$.

\input{sections/ablation}

\subsection{Runtime and the cost of consistency}
\Cref{tab:robot} (right) prices the property on the simulation grid: the
three millisecond baselines carry a point estimate and flip $26\,\%$ of
goals after elimination; B2 brings that to $10\,\%$ for $1.0\,\mathrm{s}$
per query;
\method\ reproduces the oracle at $188\,\mathrm{ms}$ median, the oracle's
own order ($162$), five times cheaper than B2, the cost being the joint draw and the conditional
walk over $64$ retained and $850$--$2395$ eliminated keyframes.

\begin{figure}[tb]
  \centering
  \includegraphics[width=\columnwidth]{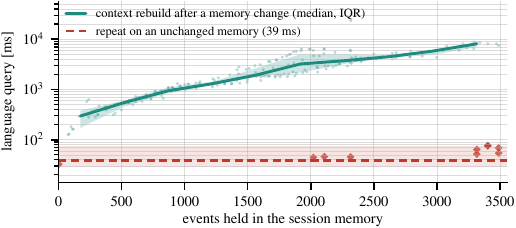}
  \caption{Language-query cost against session-memory size, on the campaign replayed on the robot (log scale). A context rebuild after a memory change grows with the events held; every further query reuses it at $39\,\mathrm{ms}$.}
  \label{fig:realtime}
\end{figure}

Replayed at sensor rate on the Jetson Thor (\cref{fig:realtime}; seven
traced runs), the memory update costs
$28\,\mathrm{ms}$ median per keyframe over $5559$ keyframes ($p_{95}$
$106$) against a $\sim\!420\,\mathrm{ms}$ budget at $2.4$ keyframes per
second, and a closure $152\,\mathrm{ms}$ median over $243$. A query splits in two:
the query-independent context is rebuilt whenever the memory changed,
linear in the events held ($0.4\,\mathrm{s}$ below $500$ events,
$4.5\,\mathrm{s}$ at $2000$--$3500$; $1.9\,\mathrm{s}$ median over $195$
rebuilds), and every further query on a settled memory costs
$39\,\mathrm{ms}$ ($28$--$52$); under a query every $10\,\mathrm{s}$ replay
throughput drops $16\,\%$. Perception takes $83$--$88\,\mathrm{ms}$ per keyframe, $36$ with TensorRT.

%% file: tables/tab_rewrites.tex
\begin{table*}[t]
\centering
\caption{Graph rewrites and query consistency on the campaign and on OpenLORIS-Scene: closures applied and vetted, keyframe correction under closure, and $\dproj^{\max}$ and goal flips over each row's $n$ queries, with four ablations and the negative control on the same draws, from the replay.}
\label{tab:campaign}\label{tab:openloris}
\small
\setlength{\tabcolsep}{1.9pt}
\begin{tabular}{l r rrr r rr r rr rrrrr}
\toprule
& & & & & & \multicolumn{2}{c}{correction [m]} & & & & \multicolumn{5}{c}{goal flips under ablation} \\
\cmidrule(lr){7-8}\cmidrule(lr){12-16}
& path & keyfr. & events & objects & applied & med. & max & $\dproj^{\max}$ & flips & $n$
& $-$Assoc & $-$Reliab & neg.\,ctrl & $-$Cond & $-$Corr \\
\midrule
\multicolumn{16}{l}{\emph{our campaign, four route geometries}} \\
compact, mixed & 421 & 3809 & 12\,117 & 3395 & 305 & 3.1 & 14.0 & $5.4{\times}10^{-16}$ & 0 & 108 & 8 & 19 & 20 & 65 & 65 \\
compact, outdoor & 315 & 2820 & 7379 & 2458 & 295 & 2.1 & 9.9 & $6.6{\times}10^{-16}$ & 0 & 108 & 10 & 34 & 14 & 48 & 54 \\
corridor & 757 & 2147 & 2290 & 1036 & 65 & 10.7 & 47.3 & $9.4{\times}10^{-14}$ & 0 & 36 & 3 & 12 & 10 & 23 & 23 \\
open area & 864 & 2459 & 2846 & 1385 & 96 & 8.1 & 25.2 & $0$ & 0 & 36 & 2 & 4 & 1 & 23 & 23 \\
\midrule
\multicolumn{16}{l}{\emph{OpenLORIS-Scene, cross-session, sessions initialized at their ground-truth start pose}} \\
corridor 1--2 & --- & 1105 & 2985 & 763 & 16 & 8.3 & 36.8 & $9.1{\times}10^{-16}$ & 0 & 25 & 1 & 2 & 4 & 12 & 11 \\
market 1--2 & --- & 1066 & 6886 & 2286 & 0 & --- & --- & $0$ & 0 & 32 & 10 & 8 & 5 & 28 & 28 \\
market 1--3 & --- & 1527 & 9764 & 3537 & 30 & 7.9 & 25.2 & $0$ & 0 & 32 & 2 & 4 & 6 & 29 & 29 \\
\midrule
\textbf{total} & & 14\,933 & 44\,267 & 14\,860 & \textbf{807} & & & --- & \textbf{0} & \textbf{377} & 36 & 83 & 60 & 228 & 233 \\
\bottomrule
\end{tabular}
\end{table*}

%% file: tables/tab_robot.tex
\begin{table}[tb]
\centering
\caption{Every backend and ablation on the campaign, from the offline rebuild of each run's graph (goal-flip rate with run-clustered 95\,\% CI and $\dproj$ against B4, ECE, MB per 1000 keyframes beyond the shared event store; $^{\dagger}$$0.08$ reduced graph $+$ $6.12$ archive, $1.6$ under a fill-reducing order we did not run) and on the simulation grid's clean condition (median latency, mean $\dproj$, flip rate over $8000$ queries per backend); a $0$ count is $\le 1.0\,\%$ at $95\,\%$.}
\label{tab:robot}\label{tab:cost}
\scriptsize
\setlength{\tabcolsep}{2.5pt}
\begin{tabular}{l l r r r  r r r}
\toprule
& \multicolumn{4}{c}{campaign (robot)} & \multicolumn{3}{c}{simulation} \\
\cmidrule(lr){2-5}\cmidrule(lr){6-8}
& flips \% [CI] & $\dproj$ & ECE & MB & ms & $\dproj$ & flips \\
\midrule
B0 world point & 18 [11, 26] & 0.071 & 0.103 & 0.13 & 1.2 & 0.157 & 0.263 \\
B1 anchor & 8 [4, 11] & 0.026 & 0.117 & 0.13 & 1.2 & 0.157 & 0.263 \\
B2 marginals & 58 [53, 62] & 0.373 & 0.121 & 0.42 & 1013 & 0.070 & 0.096 \\
B3 snapshot & 15 [10, 19] & 0.050 & 0.104 & 0.13 & 0.9 & 0.157 & 0.263 \\
B4 full graph & 0 [0, 0] & 0.000 & 0.109 & 0.67 & 162 & 0.000 & 0.000 \\
\midrule
\method & 0 [0, 0] & 0.000 & 0.109 & 6.20$^{\dagger}$ & 188 & 0.000 & 0.000 \\
\midrule
\quad $-$Association & 8 [6, 11] & 0.045 & 0.116 & --- & --- & --- & --- \\
\quad $-$Reliability & 23 [17, 29] & 0.074 & 0.122 & --- & --- & --- & --- \\
\quad negative control & 15 [9, 20] & 0.077 & 0.110 & --- & --- & --- & --- \\
\quad $-$Conditional & 56 [49, 62] & 0.363 & 0.137 & --- & --- & --- & --- \\
\quad $-$Correlation (=B2) & 58 [53, 62] & 0.373 & 0.121 & --- & --- & --- & --- \\
\bottomrule
\end{tabular}
\end{table}

%% file: tables/tab_reproj.tex
\begin{table}[tb]
\centering
\caption{Where the memory puts an object it saw once: ground-plane error at the revisit, predicted along the raw odometry (\emph{odom.}), from the birth coordinate (\emph{frozen}, B0) and from the birth measurement re-anchored on the loop-closed chain, which is \method's point prediction and B1's coordinate alike (\emph{P\,=\,B1}), against the LiDAR trajectory; medians over objects, and the fraction landing inside the detector's box at the revisit.}
\label{tab:reproj}
\scriptsize
\setlength{\tabcolsep}{2.4pt}
\begin{tabular}{lrr rrr rrr}
\toprule
& & & \multicolumn{3}{c}{error [m]} & \multicolumn{3}{c}{in box} \\
\cmidrule(lr){4-6}\cmidrule(lr){7-9}
run & objects & pairs & odom. & frozen & P\,=\,B1 & odom. & frozen & P\,=\,B1 \\
\midrule
compact, mixed 1 & 262 & 12875 & 2.25 & 2.41 & \textbf{1.09} & 0.07 & 0.00 & \textbf{0.11} \\
compact, mixed 2 & 256 & 14806 & 2.63 & 3.42 & \textbf{1.14} & 0.09 & 0.03 & \textbf{0.18} \\
compact, mixed 3 & 277 & 16869 & 1.55 & 1.72 & \textbf{1.00} & 0.07 & 0.09 & \textbf{0.12} \\
\midrule
compact, outdoor 1 & 127 & 9311 & 3.49 & 1.80 & \textbf{0.80} & 0.00 & 0.00 & \textbf{0.27} \\
compact, outdoor 2 & 136 & 6931 & 1.81 & 1.26 & \textbf{0.62} & 0.01 & 0.02 & \textbf{0.37} \\
compact, outdoor 3 & 121 & 5585 & 3.17 & 1.66 & \textbf{0.83} & 0.13 & 0.07 & \textbf{0.24} \\
\midrule
corridor, 757\,m & 243 & 9642 & 8.30 & 15.49 & \textbf{3.25} & 0.04 & 0.00 & \textbf{0.14} \\
\midrule
open area, 864\,m & 245 & 4845 & 5.38 & 9.77 & \textbf{1.71} & 0.05 & 0.01 & \textbf{0.05} \\
\midrule
\textbf{all} & 1667 & 80864 & 2.85 & 2.75 & \textbf{1.11} & 0.06 & 0.03 & \textbf{0.16} \\
\bottomrule
\end{tabular}
\end{table}

%% file: sections/ablation.tex
\subsection{Ablations and controls}
\label{sec:ablation}

The exactness test verifies posterior reconstruction; the ablations then
identify which retained structures the equivalence depends on. Each ablation removes one part, leaves the perception
cache, queries, graph, closures and draws untouched, and is evaluated on the
same 288 robot queries.
\emph{$-$Conditional} replaces the removed anchor by its marginal, the
reduction most systems make by keeping per-keyframe covariances.
\emph{$-$Correlation} replaces joint draws by independent per-variable
draws, the largest effect and one a better per-object covariance cannot
repair, the missing quantity lying between objects. \emph{$-$Reliability}
flattens event weights. \emph{$-$Association} collapses identity to its
arg-max at insertion, the smallest effect because few detections on these
routes are identity-ambiguous. \Cref{tab:robot} gives each one's rate and
$\dproj$. The \emph{negative control} permutes stored
conditionals between elimination events of equal separator dimension, so the
memory keeps as many numbers of exactly the right shapes, and the wrong ones;
without it $\dproj=0$ could mean an insensitive metric. The floor is the draw
noise: \method\ or B4 against itself under another seed reads $\dproj$
$0.001$, $1$--$2$ flips of $288$, forty times below the smallest ablation.

\paragraph{The pre-registered rule}
\method\ satisfies the first two pre-registered clauses, matching B4 in
$\dproj$ under common draws and lowering the goal-flip rate of B1--B3 after
elimination (\cref{sec:robotbase}, \cref{tab:robot}). The negative control
satisfies the third for $\dproj$ alone, leaving 15-bin calibration error and
simulated navigation success unchanged at $0.944$. That is consistent with
the control perturbing the joint distribution rather than its mean, which is
what $\dproj$ was defined to detect and what a bin-averaged calibration
error or a success rate over mean waypoints cannot.

%% file: sections/discussion.tex
\section{Discussion}
\label{sec:discussion}

\subsection{The association layer is the binding constraint}
The campaign placed three fire extinguishers as known-count objects; the
memory held $18$ and $11$ fragments of them, all near where the detector saw
them. The layer fails to merge, not to place, and the reason corrected a
documented belief: not the $\chi^2$ gate, which changes nothing when widened,
but the new-object branch of the mixture, whose implicit cosine of $1$ wins
inside the median separation of same-label fragments. Correcting that prior
merges a third of them and then saturates, the residual being stereo range error
that no association rule crosses, so effort belongs there.

\subsection{Scope of the guarantee}
Retrieval is held fixed by the shared frozen front end, so the measured gain
is consistency under graph rewrites rather than perception. The ten
navigation trials are a system-level feasibility check, the memory reaching
five objects of five and the frozen coordinate three; statistical evidence
for the representation comes from the simulation and the campaign replay,
and the negative control was not driven. A goal flip measures
rewrite-induced disagreement, not task error, and $\dproj$ deliberately
isolates that effect by comparing inference over the same factors: it is
therefore silent on whether a closure was correct and on spatial-language
reasoning, which \cref{eq:goal} does not model. B0--B3 are likewise
controlled reductions of one pipeline, isolating memory representation from
front-end differences.

The formulation targets static-object memories, with no event retirement or
reassociation for moving objects, and it bounds the live solver rather than
the historical archive, which costs six times the Bayes tree it replaces
(\cref{tab:robot}). Exact equivalence holds under the shared linearization
the campaign obtains by eliminating once after every closure; online
relinearization introduces the approximation quantified in
\cref{sec:robotbase} and reduced by periodic re-elimination, at a full-graph
solve on two closures in five. Position calibration stays determined by the
upstream sensing and graph noise models and is route-dependent
(\cref{sec:reproj}). Campaign intervals are run-clustered over eight runs
and resolve rates to about $\pm5$ points, the larger-scale evidence being
the 40-scene simulation.

%% file: sections/conclusion.tex
\section{Conclusion}

\method\ keeps semantic observations as immutable events at their birth
keyframes, retains the elimination conditional of every marginalized keyframe,
and answers a language query over the reconstructed joint posterior rather
than over committed points, $\dproj$ measuring the consistency that results. It reproduces a full-graph oracle in simulation and on a robot
campaign, elimination following the closures, where every ablation, the
negative control and a frozen coordinate flip goals it does not. Under a live bounded solver it flips $23$ of
$288$ against $53$ for a coordinate committed at insertion, and $6$ once the
conditionals are periodically re-eliminated. The results separate graph-rewrite consistency from
perception quality and quantify the approximation that online
relinearization leaves behind.

%% file: references.bib
@IEEEtranBSTCTL{IEEEexample:BSTcontrol,
  CTLuse_forced_etal       = "yes",
  CTLmax_names_forced_etal = "3",
  CTLnames_show_etal       = "1",
  CTLdash_repeated_names   = "no"
}

@inproceedings{huang2023vlmaps,
  author    = {Huang, Chenguang and Mees, Oier and Zeng, Andy and Burgard, Wolfram},
  title     = {Visual language maps for robot navigation},
  booktitle = {Proc. IEEE Int. Conf. Robot. Autom. (ICRA)},
  year      = {2023},
  pages     = {10608--10615},
  doi       = {10.1109/ICRA48891.2023.10160969}
}

@article{huang2025multimodal,
  author  = {Huang, Chenguang and Mees, Oier and Zeng, Andy and Burgard, Wolfram},
  title   = {Multimodal spatial language maps for robot navigation and manipulation},
  journal = {Int. J. Robot. Res.},
  year    = {2025},
  doi     = {10.1177/02783649251351658}
}

@article{huang2025ivlmap,
  author  = {Huang, Jiacui and Zhang, Hongtao and Zhao, Mingbo and Wu, Zhou and Liu, Yuping},
  title   = {Instance-aware visual language grounding for consumer robot navigation},
  journal = {IEEE Trans. Consum. Electron.},
  year    = {2025},
  volume  = {71},
  number  = {4},
  pages   = {12519--12526},
  doi     = {10.1109/TCE.2025.3601582}
}

@inproceedings{gu2024conceptgraphs,
  author    = {Gu, Qiao and Kuwajerwala, Ali and Morin, Sacha and Jatavallabhula, Krishna Murthy and Sen, Bipasha and Agarwal, Aditya and others},
  title     = {{ConceptGraphs}: Open-vocabulary {3D} scene graphs for perception and planning},
  booktitle = {Proc. IEEE Int. Conf. Robot. Autom. (ICRA)},
  year      = {2024},
  pages     = {5021--5028},
  doi       = {10.1109/ICRA57147.2024.10610243}
}

@inproceedings{werby2024hovsg,
  author    = {Werby, Abdelrhman and Huang, Chenguang and B{\"u}chner, Martin and Valada, Abhinav and Burgard, Wolfram},
  title     = {Hierarchical open-vocabulary {3D} scene graphs for language-grounded robot navigation},
  booktitle = {Proc. Robot.: Sci. Syst. (RSS)},
  year      = {2024},
  doi       = {10.15607/RSS.2024.XX.077}
}

@inproceedings{shafiullah2023clipfields,
  author    = {Shafiullah, Nur Muhammad Mahi and Paxton, Chris and Pinto, Lerrel and Chintala, Soumith and Szlam, Arthur},
  title     = {{CLIP-Fields}: Weakly supervised semantic fields for robotic memory},
  booktitle = {Proc. Robot.: Sci. Syst. (RSS)},
  year      = {2023},
  doi       = {10.15607/RSS.2023.XIX.074}
}

@misc{shao2025morethanapoint,
  author        = {Shao, Xinyu and Tang, Yanzhe and Xie, Pengwei and Zhou, Kaiwen and Zhuang, Yuzheng and Quan, Xingyue and others},
  title         = {More than a point: Capturing uncertainty with adaptive affordance heatmaps for spatial grounding in robotic tasks},
  year          = {2025},
  eprint        = {2510.10912},
  archivePrefix = {arXiv},
  howpublished  = {arXiv:2510.10912}
}

@misc{sitdhipol2025fplgn,
  author        = {Sitdhipol, Supawich and Sukprasongdee, Waritwong and Chuangsuwanich, Ekapol and Tse, Rina},
  title         = {Spatial language likelihood grounding network for {Bayesian} fusion of human--robot observations},
  year          = {2025},
  eprint        = {2507.19947},
  archivePrefix = {arXiv},
  howpublished  = {arXiv:2507.19947},
  note          = {To appear in Proc. IEEE Int. Conf. Syst., Man, Cybern. (SMC)}
}

@article{reijgwart2020voxgraph,
  author  = {Reijgwart, Victor and Millane, Alexander and Oleynikova, Helen and Siegwart, Roland and Cadena, Cesar and Nieto, Juan},
  title   = {Voxgraph: Globally consistent, volumetric mapping using signed distance function submaps},
  journal = {IEEE Robot. Autom. Lett.},
  year    = {2020},
  volume  = {5},
  number  = {1},
  pages   = {227--234},
  doi     = {10.1109/LRA.2019.2953859}
}

@inproceedings{hughes2022hydra,
  author    = {Hughes, Nathan and Chang, Yun and Carlone, Luca},
  title     = {Hydra: A real-time spatial perception system for {3D} scene graph construction and optimization},
  booktitle = {Proc. Robot.: Sci. Syst. (RSS)},
  year      = {2022},
  doi       = {10.15607/RSS.2022.XVIII.050}
}

@article{hughes2024foundations,
  author  = {Hughes, Nathan and Chang, Yun and Hu, Siyi and Talak, Rajat and Abdulhai, Rumaia and Strader, Jared and Carlone, Luca},
  title   = {Foundations of spatial perception for robotics: Hierarchical representations and real-time systems},
  journal = {Int. J. Robot. Res.},
  year    = {2024},
  volume  = {43},
  number  = {10},
  pages   = {1457--1505},
  doi     = {10.1177/02783649241229725}
}

@article{maggio2024clio,
  author  = {Maggio, Dominic and Chang, Yun and Hughes, Nathan and Trang, Matthew and Griffith, Dan and Dougherty, Carlyn and others},
  title   = {Clio: Real-time task-driven open-set {3D} scene graphs},
  journal = {IEEE Robot. Autom. Lett.},
  year    = {2024},
  volume  = {9},
  number  = {10},
  pages   = {8921--8928},
  doi     = {10.1109/LRA.2024.3451395}
}

@inproceedings{schmid2024khronos,
  author    = {Schmid, Lukas and Abate, Marcus and Chang, Yun and Carlone, Luca},
  title     = {Khronos: A unified approach for spatio-temporal metric-semantic {SLAM} in dynamic environments},
  booktitle = {Proc. Robot.: Sci. Syst. (RSS)},
  year      = {2024},
  doi       = {10.15607/RSS.2024.XX.081}
}

@article{kretzschmar2012compression,
  author  = {Kretzschmar, Henrik and Stachniss, Cyrill},
  title   = {Information-theoretic compression of pose graphs for laser-based {SLAM}},
  journal = {Int. J. Robot. Res.},
  year    = {2012},
  volume  = {31},
  number  = {11},
  pages   = {1219--1230},
  doi     = {10.1177/0278364912455072}
}

@article{carlevarisbianco2014glc,
  author  = {Carlevaris-Bianco, Nicholas and Kaess, Michael and Eustice, Ryan M.},
  title   = {Generic node removal for factor-graph {SLAM}},
  journal = {IEEE Trans. Robot.},
  year    = {2014},
  volume  = {30},
  number  = {6},
  pages   = {1371--1385},
  doi     = {10.1109/TRO.2014.2347571}
}

@inproceedings{mazuran2014sparsification,
  author    = {Mazuran, Mladen and Tipaldi, Gian Diego and Spinello, Luciano and Burgard, Wolfram},
  title     = {Nonlinear graph sparsification for {SLAM}},
  booktitle = {Proc. Robot.: Sci. Syst. (RSS)},
  year      = {2014},
  doi       = {10.15607/RSS.2014.X.040}
}

@article{kaess2012isam2,
  author  = {Kaess, Michael and Johannsson, Hordur and Roberts, Richard and Ila, Viorela and Leonard, John J. and Dellaert, Frank},
  title   = {{iSAM2}: Incremental smoothing and mapping using the {Bayes} tree},
  journal = {Int. J. Robot. Res.},
  year    = {2012},
  volume  = {31},
  number  = {2},
  pages   = {216--235},
  doi     = {10.1177/0278364911430419}
}

@article{dellaert2017factorgraphs,
  author  = {Dellaert, Frank and Kaess, Michael},
  title   = {Factor graphs for robot perception},
  journal = {Found. Trends Robot.},
  year    = {2017},
  volume  = {6},
  number  = {1--2},
  pages   = {1--139},
  doi     = {10.1561/2300000043}
}

@inproceedings{bowman2017probabilistic,
  author    = {Bowman, Sean L. and Atanasov, Nikolay and Daniilidis, Kostas and Pappas, George J.},
  title     = {Probabilistic data association for semantic {SLAM}},
  booktitle = {Proc. IEEE Int. Conf. Robot. Autom. (ICRA)},
  year      = {2017},
  pages     = {1722--1729},
  doi       = {10.1109/ICRA.2017.7989203}
}

@article{doherty2022dcsam,
  author  = {Doherty, Kevin J. and Lu, Ziqi and Singh, Kurran and Leonard, John J.},
  title   = {Discrete-continuous smoothing and mapping},
  journal = {IEEE Robot. Autom. Lett.},
  year    = {2022},
  volume  = {7},
  number  = {4},
  pages   = {12395--12402},
  doi     = {10.1109/LRA.2022.3216938}
}

@inproceedings{radford2021clip,
  author    = {Radford, Alec and Kim, Jong Wook and Hallacy, Chris and Ramesh, Aditya and Goh, Gabriel and Agarwal, Sandhini and others},
  title     = {Learning transferable visual models from natural language supervision},
  booktitle = {Proc. Int. Conf. Mach. Learn. (ICML)},
  year      = {2021},
  volume    = {139},
  series    = {PMLR},
  pages     = {8748--8763}
}

@inproceedings{vasu2024mobileclip,
  author    = {Vasu, Pavan Kumar Anasosalu and Pouransari, Hadi and Faghri, Fartash and Vemulapalli, Raviteja and Tuzel, Oncel},
  title     = {{MobileCLIP}: Fast image-text models through multi-modal reinforced training},
  booktitle = {Proc. IEEE/CVF Conf. Comput. Vis. Pattern Recognit. (CVPR)},
  year      = {2024},
  pages     = {15963--15974},
  doi       = {10.1109/CVPR52733.2024.01511}
}

@inproceedings{wang2025yoloe,
  author    = {Wang, Ao and Liu, Lihao and Chen, Hui and Lin, Zijia and Han, Jungong and Ding, Guiguang},
  title     = {{YOLOE}: Real-time seeing anything},
  booktitle = {Proc. IEEE/CVF Int. Conf. Comput. Vis. (ICCV)},
  year      = {2025},
  pages     = {24591--24602},
  doi       = {10.1109/ICCV51701.2025.02280}
}

@article{labbe2019rtabmap,
  author  = {Labb{\'e}, Mathieu and Michaud, Fran{\c{c}}ois},
  title   = {{RTAB-Map} as an open-source lidar and visual simultaneous localization and mapping library for large-scale and long-term online operation},
  journal = {J. Field Robot.},
  year    = {2019},
  volume  = {36},
  number  = {2},
  pages   = {416--446},
  doi     = {10.1002/rob.21831}
}

@article{xu2022fastlio2,
  author  = {Xu, Wei and Cai, Yixi and He, Dongjiao and Lin, Jiarong and Zhang, Fu},
  title   = {{FAST-LIO2}: Fast direct {LiDAR}-inertial odometry},
  journal = {IEEE Trans. Robot.},
  year    = {2022},
  volume  = {38},
  number  = {4},
  pages   = {2053--2073},
  doi     = {10.1109/TRO.2022.3141876}
}

@inproceedings{macenski2020marathon2,
  author    = {Macenski, Steve and Mart{\'i}n, Francisco and White, Ruffin and Gin{\'e}s Clavero, Jonatan},
  title     = {The {Marathon 2}: A navigation system},
  booktitle = {Proc. IEEE/RSJ Int. Conf. Intell. Robots Syst. (IROS)},
  year      = {2020},
  pages     = {2718--2725},
  doi       = {10.1109/IROS45743.2020.9341207}
}

@inproceedings{savva2019habitat,
  author    = {Savva, Manolis and Kadian, Abhishek and Maksymets, Oleksandr and Zhao, Yili and Wijmans, Erik and Jain, Bhavana and others},
  title     = {Habitat: A platform for embodied {AI} research},
  booktitle = {Proc. IEEE/CVF Int. Conf. Comput. Vis. (ICCV)},
  year      = {2019},
  pages     = {9338--9346},
  doi       = {10.1109/ICCV.2019.00943}
}

@inproceedings{ramakrishnan2021hm3d,
  author    = {Ramakrishnan, Santhosh Kumar and Gokaslan, Aaron and Wijmans, Erik and Maksymets, Oleksandr and Clegg, Alexander and Turner, John and others},
  title     = {{Habitat-Matterport 3D Dataset} ({HM3D}): 1000 large-scale {3D} environments for embodied {AI}},
  booktitle = {Proc. Neural Inf. Process. Syst. Datasets and Benchmarks Track},
  year      = {2021}
}

@inproceedings{yadav2023hm3dsem,
  author    = {Yadav, Karmesh and Ramrakhya, Ram and Ramakrishnan, Santhosh Kumar and Gervet, Theo and Turner, John and Gokaslan, Aaron and others},
  title     = {{Habitat-Matterport 3D} semantics dataset},
  booktitle = {Proc. IEEE/CVF Conf. Comput. Vis. Pattern Recognit. (CVPR)},
  year      = {2023},
  pages     = {4927--4936},
  doi       = {10.1109/CVPR52729.2023.00477}
}

@inproceedings{shi2020openloris,
  author    = {Shi, Xuesong and Li, Dongjiang and Zhao, Pengpeng and Tian, Qinbin and Tian, Yuxin and Long, Qiwei and others},
  title     = {Are we ready for service robots? {The} {OpenLORIS-Scene} datasets for lifelong {SLAM}},
  booktitle = {Proc. IEEE Int. Conf. Robot. Autom. (ICRA)},
  year      = {2020},
  pages     = {3139--3145},
  doi       = {10.1109/ICRA40945.2020.9196638}
}

@article{holm1979,
  author  = {Holm, Sture},
  title   = {A simple sequentially rejective multiple test procedure},
  journal = {Scand. J. Stat.},
  year    = {1979},
  volume  = {6},
  number  = {2},
  pages   = {65--70}
}
